\documentclass[conference]{IEEEtran}
\IEEEoverridecommandlockouts

\usepackage{cite}
\usepackage{amsmath,amssymb,amsfonts}
\usepackage{graphicx}
\usepackage{textcomp}
\usepackage[table]{xcolor}
\usepackage[skins,breakable]{tcolorbox}
\usepackage{booktabs}
\usepackage{multirow}
\usepackage{array}
\usepackage{makecell}
\usepackage{tabularx}
\usepackage{url}
\usepackage{balance}

\definecolor{findingbg}{RGB}{242,246,239}
\newtcolorbox{findingtcolorbox}{%
    enhanced,
    breakable,
    sharp corners,
    colback=findingbg,
    colframe=findingbg,
    boxrule=0pt,
    boxsep=0pt,
    left=4pt,
    right=4pt,
    top=3pt,
    bottom=3pt,
    grow sidewards by=4pt,
    before skip=0pt,
    after skip=0pt,
}
\newcommand{\findingbox}[1]{%
    \begin{findingtcolorbox}
        #1
    \end{findingtcolorbox}
}
\newcolumntype{Y}{>{\raggedright\arraybackslash}X}

\begin{document}

\title{TicTacBench: Benchmarking Timing Closure Capabilities of Coding Agents}

\author{
    \IEEEauthorblockN{Bowei Wang\IEEEauthorrefmark{1},
        Zhigang Fang\IEEEauthorrefmark{1},
        Zhijie Yang\IEEEauthorrefmark{2},
        Renzhi Chen\IEEEauthorrefmark{3},
        Shanshan Li\IEEEauthorrefmark{1}\IEEEauthorrefmark{4},
        and Lei Wang\IEEEauthorrefmark{2}\IEEEauthorrefmark{4}\thanks{\IEEEauthorrefmark{4}Corresponding authors.}}
    \IEEEauthorblockA{\IEEEauthorrefmark{1}Computer Department,
        National University of Defense Technology,
        Changsha, China\\
        \{wangbowei, fangzhigang, shanshanli\}@nudt.edu.cn}
    \IEEEauthorblockA{\IEEEauthorrefmark{2}Defense Innovation Institute,
        Academy of Military Science,
        Beijing, China\\
        yangzhijie.nudt@foxmail.com,
        leiwang@nudt.edu.cn}
    \IEEEauthorblockA{\IEEEauthorrefmark{3}Intelligent Microelectronics Center,
        Qiyuan Lab,
        Beijing, China\\
        chenrenzhi@qiyuanlab.com}
}

\maketitle

\begin{abstract}
    Recent advances in large language models (LLMs) have led to the emergence of coding agents capable of performing complex engineering tasks, including register-transfer level (RTL) design and optimization. Existing RTL benchmarks mainly evaluate functional correctness and performance, power, and area (PPA) of the generated RTL designs, leaving agents' ability for \emph{timing closure} under-evaluated. We propose TicTacBench, a benchmark specifically designed to evaluate coding agents' capabilities for RTL-level timing closure under post-place-and-route (post-PnR) evaluation. TicTacBench contains 30 diverse tasks, each provided with a suboptimal RTL design, realistic timing constraints, functional equivalence verification, and timing reports. With over 300 runs of coding agents driven by 8 frontier LLMs, we find that even the best agent can only close 53.3\% of tasks with 7.18\% area-delay product (ADP) degradation and 8.83\% energy-delay-squared product (EDDP) improvement on average. We identify common failure categories that explain why agents fail to close timing. Then we propose TicTacSkill, a new method that guides agents to follow standard timing-closure procedures and improves the Timing Closure Rate by 9\%. These results suggest that while coding agents have made significant progress in RTL design, their timing-closure capability still has substantial room for improvement. Anonymized code and data are available at \url{https://anonymous.4open.science/r/tictacbench-anon-F3C0}.
\end{abstract}


\section{Introduction}
\label{sec:introduction}

Recently, coding agents driven by large language models (LLMs) have shown practical utility in complex engineering workflows~\cite{swebench2024}.
This progress has led to growing interest in applying them to full hardware implementation workflows.
Industrial EDA tools have adopted agent-based optimization, such as Synopsys DSO.ai and Cadence Cerebrus AI Studio.
Researchers have explored LLM-based agents for RTL generation, debugging, and RTL optimization~\cite{rtlcoder2025,vitad2025,rtlrewriter2024,symrtlo2025}.

Prior studies~\cite{pinckneyRevisitingVerilogEvalYear2025,openllmrtl2025,rtlrepo2024,cvdp2025,rtlbench2025} have developed several benchmarks to evaluate functional correctness and PPA (Performance, Power, Area) of agent-generated/optimized RTL.
For example, VerilogEvalV2~\cite{pinckneyRevisitingVerilogEvalYear2025} provides specification prompts, checks the functional correctness of the generated RTL through simulation, and reports func-pass@$k$ as the evaluation metric.
Further, RTLLM 2.0~\cite{openllmrtl2025} requires generated RTL to pass logic synthesis and extracts reports, using Worst Negative Slack (WNS), cells, and power as additional PPA metrics.

\begin{figure}[t]
    \centering
    \includegraphics[width=\columnwidth]{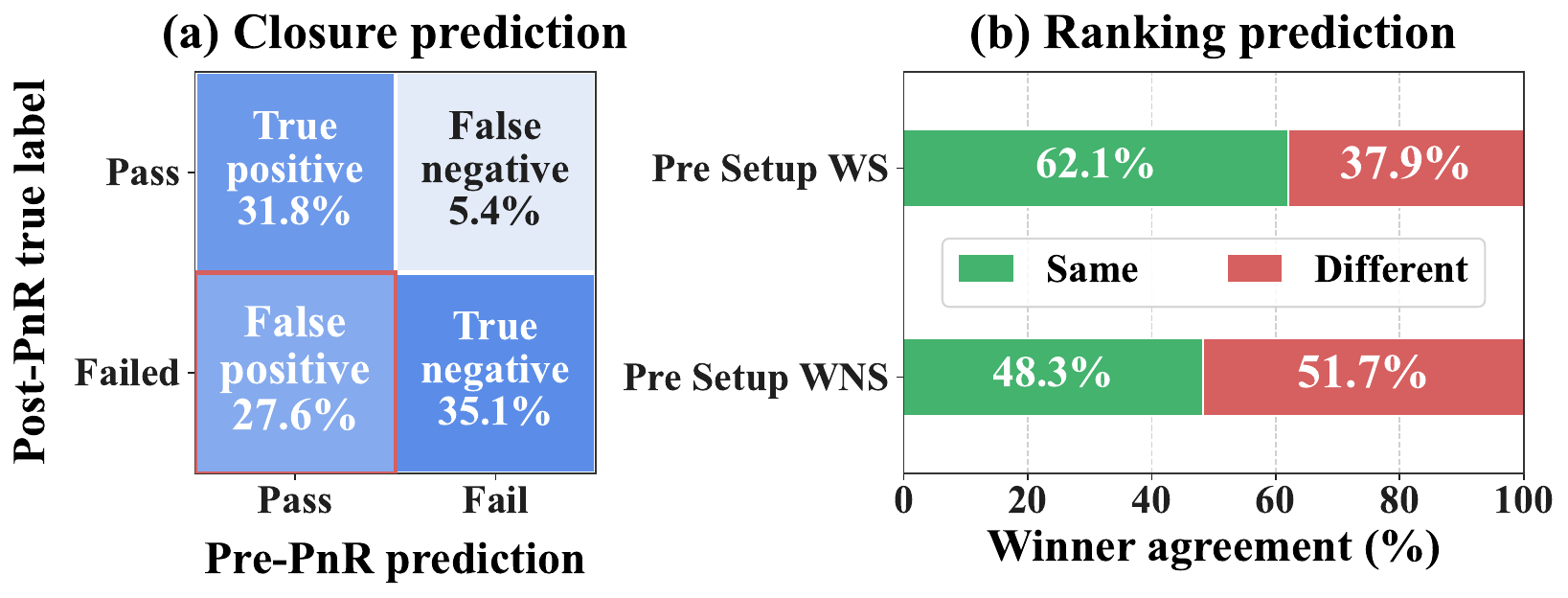}
    \caption{Pre-PnR timing metrics are unreliable for post-PnR timing closure. (a) Treating pre-PnR closure as a prediction of post-PnR closure produces high false positives: designs that appear to close timing before PnR often fail after physical implementation. (b) Pre-PnR metrics are unreliable for evaluation. The agent selected as best before PnR disagrees with the post-PnR.}
    \label{fig:intro}
\end{figure}

However, agents' capability in \emph{timing closure} remains underexplored.
Timing closure refers to the process of optimizing a hardware design to reduce critical-path negative slack and eliminate timing violations.
A design that does not close timing remains unusable even if it is functionally correct and has competitive PPA.
Existing PPA benchmarks report several pre-PnR timing metrics such as WNS and Total Negative Slack (TNS).
However, determining whether a design closes timing requires post-place-and-route Static Timing Analysis (post-PnR STA).
As Figure~\ref{fig:intro} shows, through the statistics from Section~\ref{sec:experiments}, we find that using only pre-PnR WNS to judge post-PnR timing closure leads to a high false-positive rate and incorrect result rankings.
This mismatch prevents us from gaining insight into agent timing-closure capability from existing benchmark results.

\begin{table*}[t]
    \centering
    \caption{Comparison of representative RTL benchmarks from the perspective of agent-based timing closure.}
    \label{tab:benchmarks}
    \scriptsize
    \setlength{\tabcolsep}{2pt}
    \renewcommand{\arraystretch}{0.92}
    \begin{tabular*}{\textwidth}{@{\extracolsep{\fill}}ccccc@{}}
        \toprule
        \textbf{Benchmark}                 & \textbf{Task setting}        & \textbf{Input / output}         & \textbf{Validation artifacts} & \textbf{Timing-closure coverage}     \\
        \midrule
        VerilogEval~\cite{verilogeval2023} & RTL generation               & Prompt and target RTL           & Simulation                    & No SDC or post-PnR objective         \\
        RTLLM v2~\cite{openllmrtl2025}     & RTL generation + QoR         & Spec, tests, and RTL            & Cycle-accurate                & QoR stress signal; no closure target \\
        RTLBench~\cite{rtlbench2025}        & RTL generation quality       & Spec, tests, and RTL            & Simulation, lint, LLM judge   & No timing or PPA objective           \\
        RTL-Repo~\cite{rtlrepo2024}        & Project-scale RTL generation & Repo context and Verilog        & Code checks                   & No task-level SDC or timing repair   \\
        RTLRewriter~\cite{rtlrewriter2024} & RTL rewriting for early QoR  & RTL context and rewrite         & Functional equivalence        & Logic-synthesis QoR only             \\
        RTL-OPT~\cite{rtlopt2026}          & RTL QoR optimization         & Suboptimal and optimized RTL    & Equivalence                   & PPA/QoR benchmark                    \\
        CVDP~\cite{cvdp2025}               & Realistic RTL improvement    & Project request and RTL update  & Task-specific checks          & Broad front-end tasks                \\
        \textbf{TicTacBench} (ours)        & Post-PnR timing closure      & Suboptimal design, SDC, reports & Interface, function, latency  & Post-PnR STA with timing metrics     \\
        \bottomrule
    \end{tabular*}
\end{table*}

\textbf{Research questions.}
Given this gap, we expect a dedicated evaluation to help answer the following research questions.
\textbf{RQ1:} How capable are coding agents driven by frontier LLMs at timing-closure tasks when evaluated with post-PnR STA?
\textbf{RQ2:} What failure categories explain why agents fail to close timing, indicating capability gaps?
\textbf{RQ3:} How can we improve agents' timing-closure capability beyond the baseline setting?

To address these questions, we propose \textbf{TicTacBench} (Timing Closure through Agentic Coding Benchmark), a benchmark to evaluate coding agent capabilities for RTL-level timing closure under post-PnR evaluation.
TicTacBench contains 30 diverse and challenging tasks, each provided with a suboptimal RTL design, realistic timing constraints, functional equivalence verification, pre- and post-PnR STA reports, and explicit task descriptions.
During evaluation, agents can inspect the STA reports and run the provided tool set, then submit an optimized RTL design.
To answer these research questions, we run more than 300 experiments with coding agents driven by 8 frontier LLMs on TicTacBench, collect agent trajectories, and evaluate every optimized design with post-PnR STA.
We report four metrics that capture both timing success and physical cost: Timing Closure Rate, WNS improvement, EDDP (energy-delay-squared product) improvement, and ADP (area-delay product) improvement.
We also conduct trajectory semantic analysis on the agent trajectories to identify agent capabilities and failure categories.
The results show that even the GPT-5.4-based agent can close only 53.3\% of tasks.
The optimized designs improve EDDP by 8.83\% on average, but degrade ADP by 7.18\% on average.
The trajectory semantic analysis shows that failure categories vary across agents, indicating gaps in RTL domain knowledge, RTL architecture choices, and optimization workflow capabilities.

Based on these findings, we propose \textbf{TicTacSkill}, which guides agents to use more suitable timing-closure procedures and improves the Timing Closure Rate by 9\%.
Together, these results show that coding agents still face a capability gap in industrial hardware design tasks, and point to report-grounded RTL reasoning, structural timing optimization, and flow-aware validation as promising directions for future research.

Our main contributions are as follows:

\begin{itemize}
    \item We propose TicTacBench, to our knowledge the first benchmark that makes post-PnR timing closure the primary evaluation target for coding agents. It provides a realistic and fair evaluation for measuring whether coding agents can achieve real timing closure, a capability largely absent from existing RTL benchmarks.
    \item We propose several timing-closure metrics that measure both timing success and physical cost. We evaluate coding agents driven by 8 frontier LLMs over more than 300 runs and show that even the best agent closes only half of tasks, with timing gains often accompanied by area degradation.
    \item We analyze agent trajectories and identify 6 recurring failure categories grouped into RTL domain knowledge, RTL architecture design, and optimization flow, indicating specific capability gaps.
    \item We propose TicTacSkill as a baseline method for timing closure tasks, exploring how agent capabilities can be improved for post-PnR timing closure through guidance.
\end{itemize}

\section{Benchmarking Agent-based Timing Closure}
\label{sec:prerequisite}

In this section, we first review representative RTL benchmarks, summarize their task settings and metrics, and analyze why no single existing benchmark fully covers the design requirements needed to evaluate agent-based timing-closure capability. Then, we revisit the definition and characteristics of timing closure in real implementation flow, and formalize the concept of Agent-based Timing Closure (ATC) Task.

\subsection{Inspection of Current Benchmarks}
\label{sec:inspection}

In this section, we review several representative benchmarks that are widely recognized, and summarize their task settings and evaluation metrics in Table~\ref{tab:benchmarks}.
From the perspective of the characteristics of ATC, we argue that \emph{no single existing benchmark covers the design requirements needed to evaluate ATC capability}.
This motivates us to introduce the TicTacBench.

\textbf{Lack of Timing Target/Constraints.}
Existing PPA benchmarks do not expose timing as a reproducible, design-specific timing-closure target.
Across RTLLM v2, RTL-OPT, CVDP, and RTLRewriter, the reported QoR results rely on logic-synthesis artifacts that are not released as part of the task artifacts.
RTLLM used an aggressively high frequency of $0.1$ ns so that the synthesized designs always produced WNS, turning timing into a stress signal rather than a meaningful closure objective.

\begin{figure*}[t]
    \centering
    \includegraphics[width=0.85\textwidth]{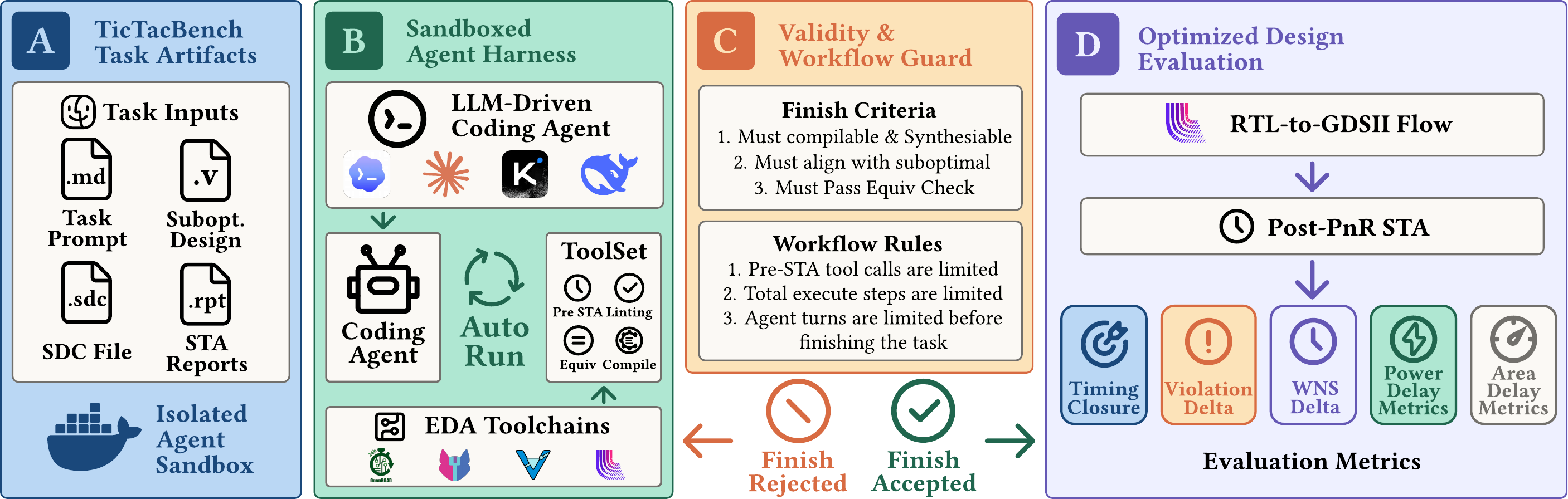}
    \caption{TicTacBench evaluation flow. The agent uses released task artifacts and tool set commands to submit \texttt{optimized.v}; valid candidates pass interface, function, and latency guards before post-PnR evaluation computes closure, WNS repair, power-delay trade-offs, and area-delay trade-offs.}
    \label{fig:method}
\end{figure*}

\textbf{Cycle-Accurate Verification Bias.}
Existing benchmarks often rely on cycle-accurate testbenches that are stricter than timing closure requires.
For example, the width\_8to16 task in RTLLM checked \texttt{valid\_out == 0} after one \texttt{\#10} step and later required \texttt{(data\_out == 16'b1010000010100001 \&\& valid\_out == 1)} at another fixed checkpoint.
As a result, cycle-accurate equivalence unnecessarily narrows the optimization space available to the agent, making latency-changing optimizations infeasible.

\textbf{Lack of Post-PnR STA Ground Truth.}
Most existing RTL benchmarks report timing outcomes only through synthesis-stage QoR metrics such as WNS, TNS, or other PPA indicators.
These metrics cannot evaluate whether an agent can use post-PnR timing feedback to drive further RTL-level repair.

\subsection{Definition and Characteristics of Timing Closure}
\label{sec:definition}

\emph{Timing closure} refers to the process of refining a hardware design to ensure the timing constraints of sequential elements, such as flip-flops, are satisfied under a target implementation environment~\cite{kahng2011}.
Failing to achieve timing closure can lead to incorrect sampling or unstable behavior at the target frequency.
The two classes of timing constraints that must be satisfied are as follows:

\emph{Setup constraint $t_{setup}$}: requires the data at the input to remain stable at least $t_{setup}$ before the active clock edge.

\emph{Hold constraint $t_{hold}$}: requires the data at the input to remain stable for at least $t_{hold}$ after the active clock edge.

Timing closure can be achieved through a variety of refinement methods, including RTL microarchitectural adjustments~\cite{retiming1991}, logic synthesis optimization~\cite{esyn2024}, clock-tree construction and placement-and-routing refinement~\cite{clocktreeaware2016}, and post-PnR engineering change order (ECO) adjustments~\cite{iraware2024}.
Microarchitectural adjustments and logic synthesis optimization are typically used to satisfy setup constraints, whereas hold repair often requires adjustments during the physical-design stage.
In this paper, we focus on enabling coding agents to resolve setup violations through RTL microarchitectural optimization, thereby achieving post-PnR setup timing closure.
Hold violations usually require physical-level adjustments, and the available interventions at the RTL level are much more limited; we therefore leave them out of scope.

We formalize an \textit{ATC (Agent-based Timing Closure) task} as $x = (r, e, c, v, q, m)$, where $r$ denotes the input RTL design; $e$ denotes the implementation environment, including technology library, flow configuration, and timing-analysis corners that affect timing analysis and optimization results; $c$ denotes the timing constraints, represented by SDC (Synopsys Design Constraints) files, which specify the target clock and timing assumptions under which $r$ initially has setup violations in $e$; $v$ denotes the equivalence verification artifacts; $q$ denotes the released timing evidence available to the agent, including baseline pre- and post-PnR reports; and $m$ denotes any other task-specific requirements.
Let $U = \{u_{\mathrm{comp}}, u_{\mathrm{equiv}}, u_{\mathrm{sta}}\}$ be the agent's minimal tool set, corresponding to compilation, equivalence verification, and pre-PnR STA, respectively.
The agent can be modeled as $A_U(x) = r'$, where the output is a new RTL design $r'$.
The goal of the task is, for each specific input instance, to start from a design $r$ with violations and produce an $r'$ that either achieves timing closure or moves measurably closer to it under the same $e$ and $c$.

\section{Benchmark Construction and Evaluation}
\label{sec:evaluation}

In this section, we describe how TicTacBench is constructed and how coding agents are evaluated.
We then describe the evaluation flow and the post-PnR metrics used to measure timing closure, WNS repair, and the physical cost of each optimized design.

\subsection{Task Construction}
\label{sec:construction_source}

We curated candidate designs from RTLLM 2.0~\cite{openllmrtl2025} and RTL-OPT~\cite{rtlopt2026} as they provide high-quality baseline RTL with optimization headroom.
We kept 25 out of 90 designs whose critical paths come from arithmetic, pipeline, or control logic.
CDC and memory/storage-centered tasks are left to future benchmark splits; small standalone modules are excluded when they offer too little timing-optimization headroom.

There are two kinds of ATC tasks in TicTacBench: \emph{latency-preserving} tasks require the optimized design to have the same latency as the baseline, while \emph{latency-changeable} tasks allow the optimized design to have explicitly required additional latency. We derived 5 latency-changeable tasks from the original tasks, encompassing total 30 tasks in the benchmark.

\begin{figure*}[t]
    \centering
    \includegraphics[width=0.9\textwidth]{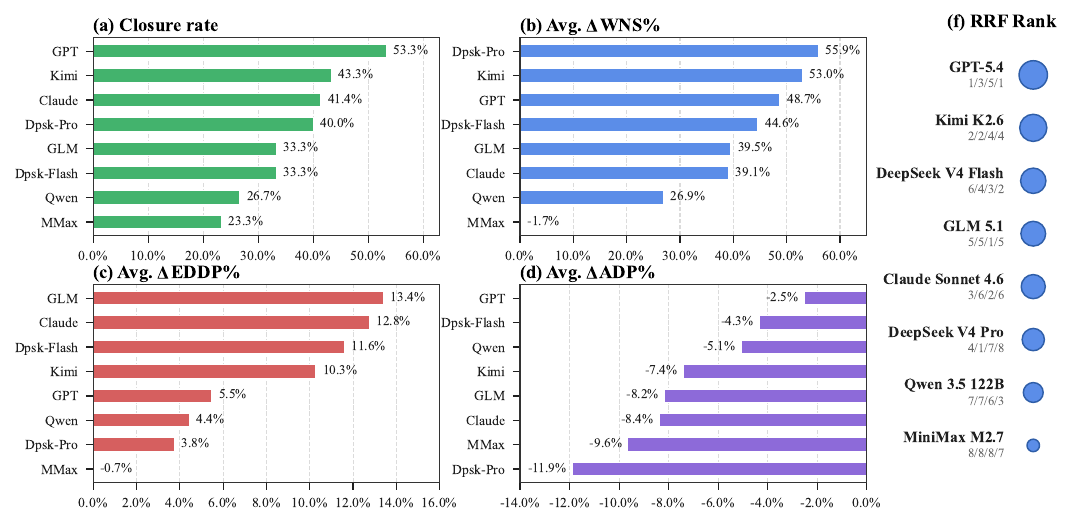}
    \caption{Main TicTacBench leaderboard from 30 tasks and coding agents driven by 8 frontier LLMs. Panel (a) ranks agents by post-PnR Timing Closure Rate. Panels (b)--(d) rank agents by Avg. $\Delta$ WNS\%, Avg. $\Delta$ EDDP\%, and Avg. $\Delta$ ADP\% relative to the suboptimal baseline; higher is better. Panel (f) reports the Reciprocal Rank Fusion (RRF) ranking across the four metrics.}
    \label{fig:main_leaderboard}
\end{figure*}

\subsection{Evaluation Flow}
\label{sec:evaluation_flow}

Figure~\ref{fig:method} illustrates the agent harness and evaluation flow.
The agent receives the task input and may inspect files, edit RTL, and run tools.
The baseline post-PnR reports describe the original failure, so the agent can localize the critical path and reason from reports.
During exploration, the agent runs in a sandboxed environment with the given minimal tool set. To keep runs resource-bounded and comparable, it may run the pre-PnR STA tool only a limited number of times, but cannot run post-PnR STA since full physical implementation has substantial resource and time cost and is usually strictly limited and scheduled in real implementation flows.
The typical agent workflow is: diagnose the baseline violation, implement a synthesizable RTL transformation, check compile and equivalence, use and interpret pre-PnR STA, and submit the best final candidate.

We integrate full open source EDA toolchains for reproducibility, including Yosys for synthesis, SymbiYosys for equivalence checking, LibreLane~\cite{openlane2020} with OpenROAD and OpenSTA for full-flow post-PnR evaluation.
For tasks without a latency allowance, the optimized design must be equivalent to the baseline under the same output timing.
For tasks with an explicit allowed additional latency, the equivalence checker aligns the allowed cycle shift and rejects any other latency changes.
After these checks pass, we rerun the full implementation flow on the optimized RTL in a clean environment.

\subsection{Evaluation Metrics}
\label{sec:evaluation_metrics}

We propose to use 4 dedicated metrics to characterize timing closure results.
We first define the effective clock period $t_{\mathrm{ECP}}$.
For task $i$, let $t^i$ be the clock period specified by the constraint, and let $t_{\mathrm{WNS}}^i$ be the post-PnR WNS, where closed designs have $t_{\mathrm{WNS}}^i = 0$.
The effective clock period $t_{\mathrm{ECP}}$ is calculated as $t_{\mathrm{ECP}}^{i} = t^i - t_{\mathrm{WNS}}^i$.
This calculation intentionally ignores the positive part of slack and therefore measures relative to the constraint rather than as unconstrained timing optimization.
In the following metrics, $sub.$ denotes the suboptimal baseline design, and $opt.$ denotes the optimized design from the agent.

The \emph{Timing Closure Rate $r$} is
\begin{equation}
    r = \frac{1}{N}\sum_{i=1}^{N}\mathbf{1}\left[t_{\mathrm{WNS}}^{opt., i} = 0\right] \times 100\%,
\end{equation}
which measures the overall fraction of benchmark tasks whose final optimized RTL is judged timing closed.

The \emph{Average WNS improvement rate} is
\begin{equation}
    \mathrm{Avg.}\ \Delta t_{\mathrm{WNS}}\% =
    \frac{1}{N}\sum_{i=1}^{N}\frac{t_{\mathrm{WNS}}^{opt.} - t_{\mathrm{WNS}}^{sub.}}{0 - t_{\mathrm{WNS}}^{sub.}}\times 100\%,
\end{equation}
which measures the average fraction of the baseline WNS improved by the optimized design.

The \emph{Average EDDP Improvement Rate} is
\begin{equation}
    \begin{aligned}
        \mathrm{Avg.}\ \Delta \mathrm{EDDP}\% & =
        \frac{1}{N}\sum_{i=1}^{N}\frac{\mathrm{EDDP}_i^{sub.} - \mathrm{EDDP}_i^{opt.}}{\mathrm{EDDP}_i^{sub.}}\times 100\%, \\
        \mathrm{s.t.}\quad \mathrm{EDDP}_i    & = E_{\mathrm{avg},i}\times (t_{\mathrm{ECP}}^{i})^{2},
    \end{aligned}
\end{equation}
which measures the average reduction in energy-delay-squared product, an energy-performance metric in power-aware computing~\cite{stan2003poweraware}.
We use this metric to characterize the \emph{timing-prioritized power-timing tradeoff}.

\begin{table*}[t]
    \centering
    \caption{Optimized post-PnR PPA metrics by task and model. Each model group reports WNS, cell count, and power. Green groups Pareto-dominate the corresponding pre-optimization result across the three metrics, while red groups are Pareto-dominated by it. WNS is in ns, power is in mW, and cells are compacted to control density. The last row reports dominate/dominated counts over all tasks.}
    \label{tab:ppa}
    \begingroup
    \scriptsize
    \setlength{\tabcolsep}{2pt}
    \renewcommand{\arraystretch}{0.9}
    \definecolor{ppadominates}{HTML}{DFF3DF}
    \definecolor{ppadominated}{HTML}{F7DDDD}
    \resizebox{0.9\textwidth}{!}{%
        \begin{tabular}{@{}l|ccc|ccc|ccc|ccc|ccc|ccc|ccc|ccc@{}}
            \toprule
            \multicolumn{1}{l}{\textbf{Task}} & \multicolumn{3}{c}{\textbf{GPT}} & \multicolumn{3}{c}{\textbf{Claude}} & \multicolumn{3}{c}{\textbf{Dpsk-Pro}} & \multicolumn{3}{c}{\textbf{Dpsk-Flash}} & \multicolumn{3}{c}{\textbf{Kimi}} & \multicolumn{3}{c}{\textbf{MiniMax}} & \multicolumn{3}{c}{\textbf{GLM}} & \multicolumn{3}{c}{\textbf{Qwen}} \\
            \cmidrule(lr){2-4}\cmidrule(lr){5-7}\cmidrule(lr){8-10}\cmidrule(lr){11-13}\cmidrule(lr){14-16}\cmidrule(lr){17-19}\cmidrule(lr){20-22}\cmidrule(lr){23-25}
            \multicolumn{1}{l}{} & \multicolumn{1}{c}{\textbf{WNS}} & \multicolumn{1}{c}{\textbf{Cells}} & \multicolumn{1}{c}{\textbf{Power}} & \multicolumn{1}{c}{\textbf{WNS}} & \multicolumn{1}{c}{\textbf{Cells}} & \multicolumn{1}{c}{\textbf{Power}} & \multicolumn{1}{c}{\textbf{WNS}} & \multicolumn{1}{c}{\textbf{Cells}} & \multicolumn{1}{c}{\textbf{Power}} & \multicolumn{1}{c}{\textbf{WNS}} & \multicolumn{1}{c}{\textbf{Cells}} & \multicolumn{1}{c}{\textbf{Power}} & \multicolumn{1}{c}{\textbf{WNS}} & \multicolumn{1}{c}{\textbf{Cells}} & \multicolumn{1}{c}{\textbf{Power}} & \multicolumn{1}{c}{\textbf{WNS}} & \multicolumn{1}{c}{\textbf{Cells}} & \multicolumn{1}{c}{\textbf{Power}} & \multicolumn{1}{c}{\textbf{WNS}} & \multicolumn{1}{c}{\textbf{Cells}} & \multicolumn{1}{c}{\textbf{Power}} & \multicolumn{1}{c}{\textbf{WNS}} & \multicolumn{1}{c}{\textbf{Cells}} & \multicolumn{1}{c}{\textbf{Power}} \\
            \midrule
            accu & +0.00 & +893 & 1.37 & -0.04 & +1.17k & 1.68 & -0.01 & +901 & 1.36 & -0.05 & +903 & 1.38 & -0.03 & +967 & 1.41 & -0.17 & +885 & 1.40 & +0.00 & +1.00k & 1.46 & \cellcolor[HTML]{DFF3DF}-0.04 & \cellcolor[HTML]{DFF3DF}+827 & \cellcolor[HTML]{DFF3DF}1.30 \\
            add\_sub & \cellcolor[HTML]{DFF3DF}-0.12 & \cellcolor[HTML]{DFF3DF}+353 & \cellcolor[HTML]{DFF3DF}0.38 & -0.07 & +599 & 0.37 & -0.08 & +592 & 0.38 & \cellcolor[HTML]{DFF3DF}-0.12 & \cellcolor[HTML]{DFF3DF}+353 & \cellcolor[HTML]{DFF3DF}0.38 & -0.06 & +367 & 0.45 & -0.10 & +611 & 0.37 & -0.06 & +367 & 0.45 & \cellcolor[HTML]{DFF3DF}-0.07 & \cellcolor[HTML]{DFF3DF}+323 & \cellcolor[HTML]{DFF3DF}0.37 \\
            adder\_32bit & \cellcolor[HTML]{F7DDDD}-0.21 & \cellcolor[HTML]{F7DDDD}+2.76k & \cellcolor[HTML]{F7DDDD}0.64 & \cellcolor[HTML]{F7DDDD}-0.23 & \cellcolor[HTML]{F7DDDD}+2.65k & \cellcolor[HTML]{F7DDDD}0.64 & \cellcolor[HTML]{F7DDDD}-0.14 & \cellcolor[HTML]{F7DDDD}+1.71k & \cellcolor[HTML]{F7DDDD}0.49 & \cellcolor[HTML]{DFF3DF}-0.07 & \cellcolor[HTML]{DFF3DF}+1.53k & \cellcolor[HTML]{DFF3DF}0.44 & \cellcolor[HTML]{DFF3DF}-0.07 & \cellcolor[HTML]{DFF3DF}+1.53k & \cellcolor[HTML]{DFF3DF}0.44 & \cellcolor[HTML]{DFF3DF}-0.07 & \cellcolor[HTML]{DFF3DF}+1.53k & \cellcolor[HTML]{DFF3DF}0.44 & \cellcolor[HTML]{F7DDDD}-0.11 & \cellcolor[HTML]{F7DDDD}+1.73k & \cellcolor[HTML]{F7DDDD}0.49 & \cellcolor[HTML]{DFF3DF}-0.07 & \cellcolor[HTML]{DFF3DF}+1.53k & \cellcolor[HTML]{DFF3DF}0.44 \\
            adder\_pipe\_64bit & -0.23 & +11.8k & 7.93 & +0.00 & +12.7k & 8.09 & -0.23 & +11.8k & 7.93 & -0.23 & +11.8k & 7.93 & -0.23 & +11.8k & 7.93 & -0.33 & +12.1k & 7.76 & -0.22 & +12.2k & 7.79 & -0.23 & +11.8k & 7.93 \\
            adder\_select & +0.00 & +1.29k & 0.50 & +0.00 & +1.29k & 0.50 & +0.00 & +1.29k & 0.50 & +0.00 & +1.58k & 0.60 & +0.00 & +1.29k & 0.50 & +0.00 & +1.29k & 0.50 & +0.00 & +1.29k & 0.50 & +0.00 & +1.29k & 0.50 \\
            addr\_calcu & -0.05 & +1.14k & 0.55 & -0.07 & +888 & 0.45 & -0.09 & +1.37k & 0.51 & \cellcolor[HTML]{DFF3DF}-0.04 & \cellcolor[HTML]{DFF3DF}+813 & \cellcolor[HTML]{DFF3DF}0.37 & -0.02 & +928 & 0.59 & -0.23 & +839 & 0.64 & -0.09 & +1.35k & 0.44 & -0.02 & +928 & 0.59 \\
            addr\_calcu\_pipe & +0.00 & +1.19k & 0.70 & -0.05 & +1.09k & 0.69 & -0.08 & +2.14k & 0.94 & -0.08 & +2.04k & 0.73 & -0.10 & +1.61k & 0.65 & +0.00 & +2.34k & 1.13 & -0.02 & +2.35k & 1.04 & +0.00 & +1.90k & 1.23 \\
            alu & -1.05 & +5.62k & 0.47 & -0.99 & +4.76k & 0.44 & +0.00 & +5.15k & 0.56 & -0.29 & +5.58k & 0.47 & +0.00 & +5.16k & 0.60 & -0.29 & +5.58k & 0.47 & +0.00 & +5.19k & 0.59 & \cellcolor[HTML]{F7DDDD}-1.37 & \cellcolor[HTML]{F7DDDD}+6.08k & \cellcolor[HTML]{F7DDDD}0.68 \\
            alu\_64bit & \cellcolor[HTML]{DFF3DF}+0.00 & \cellcolor[HTML]{DFF3DF}+3.69k & \cellcolor[HTML]{DFF3DF}0.68 & \cellcolor[HTML]{F7DDDD}-0.49 & \cellcolor[HTML]{F7DDDD}+5.50k & \cellcolor[HTML]{F7DDDD}0.83 & \cellcolor[HTML]{DFF3DF}+0.00 & \cellcolor[HTML]{DFF3DF}+4.77k & \cellcolor[HTML]{DFF3DF}0.71 & \cellcolor[HTML]{DFF3DF}+0.00 & \cellcolor[HTML]{DFF3DF}+3.89k & \cellcolor[HTML]{DFF3DF}0.59 & \cellcolor[HTML]{DFF3DF}+0.00 & \cellcolor[HTML]{DFF3DF}+3.43k & \cellcolor[HTML]{DFF3DF}0.65 & +0.00 & +5.01k & 0.72 & -0.13 & +5.13k & 0.81 & +0.00 & +4.86k & 0.71 \\
            calculation & \cellcolor[HTML]{DFF3DF}+0.00 & \cellcolor[HTML]{DFF3DF}+1.91k & \cellcolor[HTML]{DFF3DF}0.65 & \cellcolor[HTML]{DFF3DF}+0.00 & \cellcolor[HTML]{DFF3DF}+2.02k & \cellcolor[HTML]{DFF3DF}0.55 & +0.00 & +2.26k & 0.63 & +0.00 & +2.14k & 0.66 & +0.00 & +2.09k & 0.66 & +0.00 & +2.02k & 0.55 & \cellcolor[HTML]{DFF3DF}+0.00 & \cellcolor[HTML]{DFF3DF}+2.01k & \cellcolor[HTML]{DFF3DF}0.64 & +0.00 & +2.03k & 0.61 \\
            calculation\_pipe & +0.00 & +3.39k & 1.38 & +0.00 & +3.21k & 0.83 & +0.00 & +3.32k & 0.97 & +0.00 & +2.77k & 0.95 & +0.00 & +3.35k & 0.93 & +0.00 & +3.21k & 0.88 & +0.00 & +3.22k & 0.81 & +0.00 & +3.57k & 1.08 \\
            divider\_16bit & \cellcolor[HTML]{DFF3DF}+0.00 & \cellcolor[HTML]{DFF3DF}+7.03k & \cellcolor[HTML]{DFF3DF}11.2 & -0.18 & +10.3k & 246 & -0.03 & +10.3k & 250 & -0.18 & +10.3k & 246 & +0.00 & +10.3k & 245 & -0.18 & +10.3k & 246 & -0.18 & +10.3k & 246 & -0.18 & +10.3k & 246 \\
            divider\_32bit & \cellcolor[HTML]{DFF3DF}+0.00 & \cellcolor[HTML]{DFF3DF}+27.4k & \cellcolor[HTML]{DFF3DF}94.5 & \cellcolor[HTML]{DFF3DF}+0.00 & \cellcolor[HTML]{DFF3DF}+35.5k & \cellcolor[HTML]{DFF3DF}12.4 & +0.00 & +68.9k & 306 & \cellcolor[HTML]{DFF3DF}+0.00 & \cellcolor[HTML]{DFF3DF}+35.6k & \cellcolor[HTML]{DFF3DF}12.1 & \cellcolor[HTML]{DFF3DF}+0.00 & \cellcolor[HTML]{DFF3DF}+30.3k & \cellcolor[HTML]{DFF3DF}209 & \cellcolor[HTML]{DFF3DF}+0.00 & \cellcolor[HTML]{DFF3DF}+27.8k & \cellcolor[HTML]{DFF3DF}57.1 & \cellcolor[HTML]{DFF3DF}+0.00 & \cellcolor[HTML]{DFF3DF}+26.5k & \cellcolor[HTML]{DFF3DF}53.0 & -0.22 & +49.6k & 213 \\
            divider\_32bit\_pipe & \cellcolor[HTML]{DFF3DF}+0.00 & \cellcolor[HTML]{DFF3DF}+44.2k & \cellcolor[HTML]{DFF3DF}8.23 & +0.00 & +57.1k & 63.3 & -1.40 & +56.6k & 71.1 & -1.78 & +57.8k & 77.0 & -1.40 & +56.6k & 71.1 & -2.67 & +52.3k & 296 & -1.40 & +56.6k & 71.1 & +0.00 & +57.0k & 35.0 \\
            fixed\_point\_adder & +0.00 & +1.52k & 0.63 & -0.20 & +1.52k & 0.64 & \cellcolor[HTML]{F7DDDD}-0.22 & \cellcolor[HTML]{F7DDDD}+1.55k & \cellcolor[HTML]{F7DDDD}0.61 & +0.00 & +1.52k & 0.61 & \cellcolor[HTML]{F7DDDD}-0.51 & \cellcolor[HTML]{F7DDDD}+1.84k & \cellcolor[HTML]{F7DDDD}0.66 & \cellcolor[HTML]{F7DDDD}-0.29 & \cellcolor[HTML]{F7DDDD}+1.54k & \cellcolor[HTML]{F7DDDD}0.60 & \cellcolor[HTML]{F7DDDD}-0.37 & \cellcolor[HTML]{F7DDDD}+1.55k & \cellcolor[HTML]{F7DDDD}0.61 & \cellcolor[HTML]{F7DDDD}-0.22 & \cellcolor[HTML]{F7DDDD}+1.55k & \cellcolor[HTML]{F7DDDD}0.61 \\
            float\_multi & \cellcolor[HTML]{DFF3DF}-0.22 & \cellcolor[HTML]{DFF3DF}+17.1k & \cellcolor[HTML]{DFF3DF}7.87 & \cellcolor[HTML]{DFF3DF}+0.00 & \cellcolor[HTML]{DFF3DF}+17.4k & \cellcolor[HTML]{DFF3DF}7.79 & \cellcolor[HTML]{DFF3DF}+0.00 & \cellcolor[HTML]{DFF3DF}+17.5k & \cellcolor[HTML]{DFF3DF}7.96 & -0.32 & +17.5k & 8.46 & -0.01 & +17.7k & 8.20 & -0.39 & +17.5k & 7.84 & \cellcolor[HTML]{F7DDDD}-0.78 & \cellcolor[HTML]{F7DDDD}+18.2k & \cellcolor[HTML]{F7DDDD}9.00 & -0.32 & +17.5k & 8.46 \\
            float\_multi\_pipe & +0.00 & +19.6k & 10.1 & +0.00 & +20.8k & 9.12 & +0.00 & +17.7k & 8.80 & -0.02 & +17.6k & 7.58 & \cellcolor[HTML]{F7DDDD}-0.88 & \cellcolor[HTML]{F7DDDD}+18.1k & \cellcolor[HTML]{F7DDDD}8.69 & +0.00 & +18.5k & 7.91 & +0.00 & +20.4k & 9.42 & -0.31 & +17.6k & 7.35 \\
            fsm & -0.15 & +552 & 0.36 & \cellcolor[HTML]{F7DDDD}-0.18 & \cellcolor[HTML]{F7DDDD}+606 & \cellcolor[HTML]{F7DDDD}0.37 & -0.13 & +591 & 0.37 & -0.15 & +552 & 0.36 & -0.04 & +593 & 0.37 & \cellcolor[HTML]{F7DDDD}-0.71 & \cellcolor[HTML]{F7DDDD}+873 & \cellcolor[HTML]{F7DDDD}0.90 & -0.11 & +549 & 0.37 & -0.15 & +552 & 0.36 \\
            fsm\_encode & \cellcolor[HTML]{DFF3DF}+0.00 & \cellcolor[HTML]{DFF3DF}+1.14k & \cellcolor[HTML]{DFF3DF}1.33 & \cellcolor[HTML]{DFF3DF}-0.10 & \cellcolor[HTML]{DFF3DF}+1.73k & \cellcolor[HTML]{DFF3DF}2.05 & -0.19 & +1.88k & 2.17 & -0.08 & +2.01k & 2.41 & \cellcolor[HTML]{DFF3DF}+0.00 & \cellcolor[HTML]{DFF3DF}+1.13k & \cellcolor[HTML]{DFF3DF}0.92 & \cellcolor[HTML]{F7DDDD}-0.44 & \cellcolor[HTML]{F7DDDD}+1.73k & \cellcolor[HTML]{F7DDDD}2.35 & -0.10 & +1.85k & 0.78 & \cellcolor[HTML]{DFF3DF}+0.00 & \cellcolor[HTML]{DFF3DF}+721 & \cellcolor[HTML]{DFF3DF}1.08 \\
            gray & +0.00 & +425 & 0.92 & +0.00 & +542 & 0.43 & +0.00 & +532 & 0.45 & +0.00 & +434 & 0.74 & +0.00 & +535 & 0.43 & -0.23 & +419 & 0.85 & +0.00 & +536 & 0.45 & +0.00 & +486 & 0.77 \\
            mac & -0.23 & +5.78k & 2.38 & -- & -- & -- & -0.20 & +5.77k & 2.42 & -0.20 & +5.77k & 2.42 & -0.30 & +5.65k & 2.43 & -0.18 & +5.19k & 3.19 & -0.23 & +5.78k & 2.38 & -0.18 & +5.19k & 3.19 \\
            mac\_pipe & -0.13 & +5.52k & 2.52 & -0.24 & +5.53k & 2.51 & -0.13 & +5.52k & 2.52 & -0.13 & +5.52k & 2.52 & -0.13 & +5.52k & 2.52 & -0.08 & +5.25k & 2.64 & -0.13 & +5.52k & 2.52 & -0.13 & +5.52k & 2.52 \\
            mul\_subexpression & -0.28 & +1.17k & 0.52 & \cellcolor[HTML]{F7DDDD}-0.27 & \cellcolor[HTML]{F7DDDD}+1.17k & \cellcolor[HTML]{F7DDDD}0.68 & -0.27 & +1.16k & 0.60 & +0.00 & +1.19k & 0.57 & \cellcolor[HTML]{F7DDDD}-0.27 & \cellcolor[HTML]{F7DDDD}+1.17k & \cellcolor[HTML]{F7DDDD}0.68 & \cellcolor[HTML]{F7DDDD}-0.27 & \cellcolor[HTML]{F7DDDD}+1.17k & \cellcolor[HTML]{F7DDDD}0.68 & \cellcolor[HTML]{F7DDDD}-0.27 & \cellcolor[HTML]{F7DDDD}+1.17k & \cellcolor[HTML]{F7DDDD}0.68 & \cellcolor[HTML]{F7DDDD}-0.27 & \cellcolor[HTML]{F7DDDD}+1.17k & \cellcolor[HTML]{F7DDDD}0.68 \\
            multi\_16bit & \cellcolor[HTML]{DFF3DF}+0.00 & \cellcolor[HTML]{DFF3DF}+1.76k & \cellcolor[HTML]{DFF3DF}1.73 & -0.14 & +2.07k & 2.08 & \cellcolor[HTML]{DFF3DF}+0.00 & \cellcolor[HTML]{DFF3DF}+1.88k & \cellcolor[HTML]{DFF3DF}1.61 & \cellcolor[HTML]{DFF3DF}+0.00 & \cellcolor[HTML]{DFF3DF}+1.78k & \cellcolor[HTML]{DFF3DF}1.68 & \cellcolor[HTML]{DFF3DF}+0.00 & \cellcolor[HTML]{DFF3DF}+1.79k & \cellcolor[HTML]{DFF3DF}1.63 & \cellcolor[HTML]{DFF3DF}-0.35 & \cellcolor[HTML]{DFF3DF}+1.99k & \cellcolor[HTML]{DFF3DF}1.93 & -0.34 & +2.14k & 1.58 & -0.42 & +2.06k & 1.96 \\
            multi\_pipe\_4bit & +0.00 & +361 & 0.95 & +0.00 & +318 & 0.95 & -0.15 & +347 & 1.06 & +0.00 & +319 & 0.96 & +0.00 & +330 & 0.95 & +0.00 & +305 & 1.00 & +0.00 & +321 & 0.95 & -0.15 & +347 & 1.06 \\
            multi\_pipe\_8bit & +0.00 & +2.01k & 2.06 & -0.34 & +1.88k & 2.15 & -0.34 & +1.88k & 2.15 & -0.34 & +1.88k & 2.15 & -0.37 & +1.87k & 2.20 & -1.73 & +1.81k & 2.15 & -0.79 & +2.02k & 2.14 & -0.51 & +1.87k & 2.12 \\
            radix2\_div & -0.15 & +2.49k & 0.87 & -0.12 & +2.39k & 0.87 & -0.12 & +2.39k & 0.88 & -0.12 & +2.38k & 0.89 & -0.19 & +2.38k & 0.86 & -0.19 & +2.38k & 0.86 & -0.19 & +2.38k & 0.86 & -0.19 & +2.38k & 0.86 \\
            serial2parallel & -0.05 & +807 & 1.11 & \cellcolor[HTML]{DFF3DF}+0.00 & \cellcolor[HTML]{DFF3DF}+679 & \cellcolor[HTML]{DFF3DF}1.05 & +0.00 & +727 & 1.12 & \cellcolor[HTML]{DFF3DF}+0.00 & \cellcolor[HTML]{DFF3DF}+719 & \cellcolor[HTML]{DFF3DF}1.06 & +0.00 & +727 & 1.12 & \cellcolor[HTML]{DFF3DF}+0.00 & \cellcolor[HTML]{DFF3DF}+679 & \cellcolor[HTML]{DFF3DF}0.98 & +0.00 & +740 & 1.32 & \cellcolor[HTML]{DFF3DF}-0.01 & \cellcolor[HTML]{DFF3DF}+677 & \cellcolor[HTML]{DFF3DF}1.05 \\
            traffic\_light & -0.06 & +975 & 0.54 & +0.00 & +822 & 0.55 & +0.00 & +680 & 1.22 & \cellcolor[HTML]{DFF3DF}+0.00 & \cellcolor[HTML]{DFF3DF}+665 & \cellcolor[HTML]{DFF3DF}0.49 & +0.00 & +986 & 0.59 & -0.04 & +681 & 1.37 & \cellcolor[HTML]{DFF3DF}+0.00 & \cellcolor[HTML]{DFF3DF}+229 & \cellcolor[HTML]{DFF3DF}0.30 & -0.23 & +738 & 0.51 \\
            width\_8to16 & -0.22 & +921 & 2.85 & -0.24 & +981 & 2.25 & -0.42 & +851 & 2.46 & -0.26 & +817 & 2.58 & -0.01 & +1.94k & 4.53 & -0.26 & +817 & 2.58 & -0.21 & +873 & 1.79 & -0.26 & +817 & 2.58 \\
            \midrule
            \multicolumn{1}{l}{\textbf{Dom./Dmd.}} & \multicolumn{3}{c}{18/3} & \multicolumn{3}{c}{5/4} & \multicolumn{3}{c}{3/2} & \multicolumn{3}{c}{17/3} & \multicolumn{3}{c}{5/3} & \multicolumn{3}{c}{14/7} & \multicolumn{3}{c}{3/4} & \multicolumn{3}{c}{5/3} \\
            \bottomrule
        \end{tabular}
    }
    \endgroup
\end{table*}

The \emph{Average ADP Improvement Rate} is
\begin{equation}
    \begin{aligned}
        \mathrm{Avg.}\ \Delta \mathrm{ADP}\% & =
        \frac{1}{N}\sum_{i=1}^{N}\frac{\mathrm{ADP}_i^{sub.} - \mathrm{ADP}_i^{opt.}}{\mathrm{ADP}_i^{sub.}}\times 100\%, \\
        \mathrm{s.t.}\quad \mathrm{ADP}_i    & = A_i\times t_{\mathrm{ECP}}^{i},
    \end{aligned}
\end{equation}
which measures the average reduction in area-delay product, an area-timing metric in logic synthesis~\cite{invictus2023}.
We use this metric to characterize the \emph{area-timing tradeoff}.

\section{Experiments}
\label{sec:experiments}

\subsection{Experimental Setup}
\label{sec:experiments_setup}

We evaluate coding agents driven by 8 frontier LLMs, including GPT-5.4~\cite{gpt54_2026}, Claude Sonnet 4.6~\cite{claude_sonnet46_2026}, Kimi K2.6~\cite{kimi_k26_2026}, DeepSeek V4 Pro and Flash~\cite{deepseek_v4_2026}, MiniMax M2.7~\cite{minimax_m27_2026}, GLM 5.1 FP8~\cite{glm51_fp8_2026}, and Qwen 3.5 122B~\cite{qwen35_122b_2026}, on the 30 TicTacBench tasks using the evaluation setup in Section~\ref{sec:evaluation}.
All reported closure, WNS, area, power, ADP, and EDDP values are computed from the post-PnR evaluation. The evaluation flow is fully automated, reproducible, and open-sourced.

\subsection{Overall Closure and QoR Results}
\label{sec:experiments_overall}

Figure~\ref{fig:main_leaderboard} gives the main leaderboard.
The strongest agent by Timing Closure Rate is the GPT-5.4-based agent, which closes 53.3\% of tasks, while the DeepSeek V4 Pro-based agent has the largest Avg. $\Delta$ WNS\%.
The next group includes the Kimi K2.6-, Claude Sonnet 4.6-, and DeepSeek V4 Pro-based agents, with Timing Closure Rates around 40\%.
The remaining agents close fewer tasks, and the MiniMax M2.7-based agent is the only one whose Avg. $\Delta$ WNS\% is slightly negative.
These results show that even state-of-the-art agents close only about half of TicTacBench at best; 23\% of tasks are not closed by any agent, and the optimizations from all agents almost always incur substantial area degradation.

The gap between Timing Closure Rate and Avg. $\Delta$ WNS\% shows why both timing metrics are needed.
The GPT-5.4-based agent has the highest Timing Closure Rate, but its Avg. $\Delta$ WNS\% (48.7\%) is lower than that of the DeepSeek V4 Pro-based agent (55.9\%) and the Kimi K2.6-based agent (53.0\%).
This result is consistent with the GPT-5.4-based agent making larger RTL rewrites: such rewrites can remove the original bottleneck and close timing, but they can also introduce new critical paths on some tasks.
The DeepSeek V4 Pro-based agent shows a different pattern.
It obtains the best Avg. $\Delta$ WNS\% but closes only 40.0\% of tasks, suggesting that local repairs often improve slack but do not always change the design enough to meet the clock constraint.

The EDDP and ADP metrics further show the PPA cost of timing closure.
Most agents improve EDDP, led by the GLM 5.1 FP8-based agent at 13.4\%, but every agent degrades ADP, from -2.5\% for the GPT-5.4-based agent to -11.9\% for the DeepSeek V4 Pro-based agent.
This indicates that current agents primarily obtain timing improvement by spending area.
By contrast, reducing delay through power-related mechanisms is harder, because it often depends on lower switching activity, clock-tree optimization, or voltage-frequency choices rather than only local RTL rewriting.

Table~\ref{tab:ppa} reports the PPA metrics, i.e., WNS, cell count, and power of each agent on each design task.
The table also counts whether each optimized result Pareto-dominates, or is dominated by, the corresponding suboptimal result across all three PPA metrics.
The GPT-5.4-based agent has the most dominating outcomes (18/30), followed by the DeepSeek V4 Flash-based agent (17/30), while the MiniMax M2.7-based agent has the most dominated outcomes (7/30).
These counts show that LLM-driven agents are not consistently Pareto-improving optimizers: stronger agents can find favorable timing-area-power trade-offs more often, but aggressive timing repairs can also produce PPA regressions.
Thus, current agents show partial ability to navigate PPA trade-offs, but they still lack stable control over when a timing improvement justifies the associated area or power cost.

\begin{figure}[t]
    \centering
    \includegraphics[width=0.9\columnwidth]{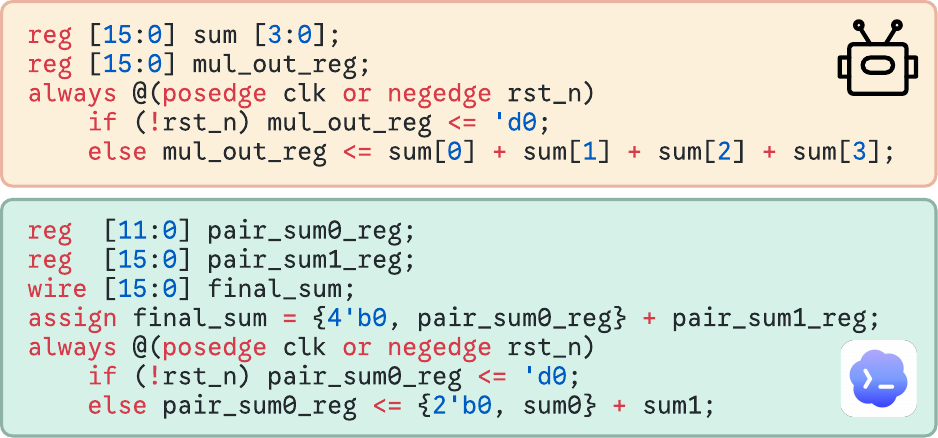}
    \caption{Representative case from multi\_pipe\_8bit.
        The failed rewrite leaves a wide addition cone on the critical stage.
        The successful rewrite moves the register boundary into the reduction tree.
    }
    \label{fig:case_study}
\end{figure}

\subsection{Case Study}
\label{sec:experiments_case_study}

Figure~\ref{fig:case_study} shows a representative case from multi\_pipe\_8bit, where the GPT-5.4-based agent is the only agent that closes post-PnR timing.
The failed solution keeps the original final reduction shape: four 16-bit partial sums are added in one registered stage before driving the output.
This leaves a wide four-input addition cone on the critical stage, and the worst failed retained run has -1.73 ns post-PnR WNS with 35 setup violations.

In contrast, the successful run performs a retiming instead of simply appending a pipeline stage.
It replaces the original output-side reduction register with two pair-sum registers, then computes the final sum from those registered pair sums.
The transformation shortens the critical stage from a four-input 16-bit reduction to narrower pair additions followed by one final add, which is the structural change needed for timing closure.

\findingbox{
    \textbf{Findings of RQ1:} We evaluated timing closure with metrics of Timing Closure Rate, WNS improvement, EDDP improvement, and ADP improvement.
    Even the best agent can close only 53.3\% of tasks.
    Successful optimizations perform micro-architectural changes, such as retiming, pipeline-boundary movement, and narrower arithmetic/reduction cones.
}

\section{Capabilities Analysis}
\label{sec:analysis}

This section analyzes the agent trajectories produced during ATC tasks to characterize agent capability beyond final metrics.
We first introduce the trajectory semantic analysis method used in this study, then aggregate its results to measure capability boundaries and failure distributions, leading to three findings about current agents' ATC capability.

\subsection{Trajectory Semantic Analysis}
\label{sec:analysis_method}

Trajectory semantic analysis treats an agent execution trajectory as behavioral evidence: the trajectory records how the agent reasons, acts through tools, observes feedback, and updates its subsequent actions.
It is especially useful for analyzing the ATC task process because an agent must read STA evidence, map the path back to RTL, choose a structural rewrite, validate equivalence, and decide whether the candidate has enough margin for physical implementation.
The final post-PnR score observes only the outcome of this process.

To apply trajectory semantic analysis, we export and preserve 240 complete agent trajectories.
Each trajectory contains the task input and final output, tool calls and tool results, and the agent's reasoning traces during the run.
We apply trajectory semantic analysis using a lightweight analyst LLM, specifically DeepSeek V4 Flash, to analyze the trajectories.
We further compare runs horizontally across agents on the same task.
For each failed run, we pair it with the same-task successful run that has the strongest post-PnR WNS.
We then identify the semantic differences between trajectories that converge successfully and those that fail to converge, cluster the differences, and obtain the 6 major failure categories.

\subsection{Failure Mechanisms and Findings}
\label{sec:analysis_findings}

Table~\ref{tab:capability} summarizes the taxonomy from the trajectory semantic analysis.
The 6 failure categories are grouped into three capabilities: RTL domain knowledge, RTL architecture design, and optimization flow.

\begin{table}[t]
    \centering
    \caption{Capability and Failure Categories of Trajectory Semantic Analysis.}
    \label{tab:capability}
    \scriptsize
    \setlength{\tabcolsep}{4pt}
    \begin{tabular}{@{}>{\raggedright\arraybackslash}m{0.14\columnwidth}>{\raggedright\arraybackslash}m{0.30\columnwidth}>{\raggedright\arraybackslash}m{0.48\columnwidth}@{}}
        \toprule
        \textbf{Capability}                   & \textbf{Failure category}                  & \textbf{Meaning}                                                                                                   \\
        \midrule
        \multirow{2}{=}{Domain\\Knowledge}    & Inefficient STA report-driven optimization & Fails to map key STA information such as slack and critical paths to the RTL level for report-driven optimization. \\
        \cmidrule(l){2-3}
                                              & Suboptimal datapath design                 & Preserves excessively deep arithmetic/datapath logic in the design.                                                \\
        \midrule
        \multirow{2}{=}{Architecture\\Design} & Suboptimal pipeline design                 & Fails to partition pipeline stages to cut the critical combinational logic chain, or introduces a new one.         \\
        \cmidrule(l){2-3}
                                              & Suboptimal data/control-path design        & Introduces control-path logic such as decode or mux logic on critical datapath paths.                              \\
        \midrule
        \multirow{2}{=}{Optimization Flow}    & Unsound exploration                        & Explores only locally suboptimal solutions and lacks a plan for exploring globally optimal solutions.              \\
        \cmidrule(l){2-3}
                                              & Unsound acceptance                         & Prematurely accepts weakly converged results from pre-PnR STA, leading to failure at post-PnR.                     \\
        \bottomrule
    \end{tabular}
\end{table}

\begin{figure}[t]
    \centering
    \includegraphics[width=0.85\columnwidth]{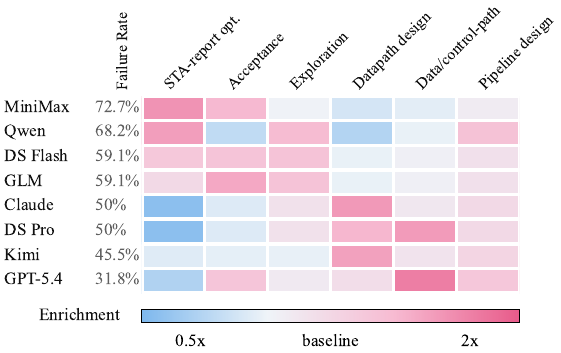}
    \caption{Failure mechanism profiles by failed trajectories. The heatmap reports each agent's failure-mechanism enrichment relative to the all-agent baseline.}
    \label{fig:capability}
\end{figure}

\textbf{Finding 1: report-driven optimization strongly affects timing-closure outcomes.}
Figure~\ref{fig:capability} shows that lower-closure agents generally fail more often on STA report-driven optimization, while higher-closure agents fail less on this dimension.
This pattern suggests that the ability to turn STA reports into targeted RTL changes has a strong effect on final optimization success.
The failed trajectories often begin by reading substantial timing-report content, but the subsequent RTL edits are not derived from the reported critical paths, cells, or slack trends.
Instead, the agent falls back to common RTL optimization templates and repeats ineffective cycles of RTL editing and STA analysis.
Successful trajectories, by contrast, convert report evidence into targeted RTL changes.
For example, in accu, the successful GPT-5.4-based agent run mapped the \texttt{\_148\_/Q $\rightarrow$ \_155\_/D} endpoint path to the accumulator carry/mux structure and rewrote that structure.
Failed peers made plausible state or adder edits, but those edits did not shorten the same post-PnR path.

\textbf{Finding 2: agents often fail to move from local RTL edits to architectural optimization.}
Many failed trajectories make syntactically valid and functionally equivalent RTL changes, but the edits remain close to the original design and do not change the microarchitecture seen by post-PnR STA.
This appears in 46.9\% of the failures, where the failed run misses at least one architectural RTL change involving datapath topology, pipeline partitioning, or control/decode placement.
Successful trajectories usually make a larger rewrite: ripple or carry-chain logic becomes a Kogge-Stone prefix adder, variable-shift logic becomes sequential shift-and-add, a wide compare or subtract is narrowed, zero flags are computed inside case branches before the result mux, or intermediate pipeline registers split a deep accumulation stage.
Failed trajectories instead stay with local optimizations, such as rewriting the \texttt{+} operator as separate assignments, trying similar adder variants, changing case/assign style, or adding a register that does not cut the actual critical path.
In adder\_pipe\_64bit, failed runs repeatedly explored similar adder variants and accepted the built-in plus operator as locally best under pre-PnR checks.
The successful run instead pre-registered generate/propagate values and used a Kogge-Stone prefix structure.
These examples show that repeated STA runs are not enough unless the agent can leave a local edit pattern and choose an architecturally stronger RTL design.

\textbf{Finding 3: different agents fail at different stages of timing closure.}
Figure~\ref{fig:capability} shows that the failure categories identify which capability is missing in each agent.
For lower-closure agents, the main bottleneck is still the basic optimization loop.
In the MiniMax M2.7, Qwen 3.5 122B, DeepSeek V4 Flash, and GLM 5.1 FP8 based agents, 76.9\%--81.25\% of failures fall into Inefficient STA report-driven optimization, Unsound exploration, or Unsound acceptance.
These runs may read reports and produce valid RTL, but they do not reliably choose the right RTL change, compare enough alternatives, or reject a candidate that is too weak for post-PnR closure.
For higher-closure agents, the remaining bottleneck is more often the RTL design choice itself: in the Claude Sonnet 4.6, DeepSeek V4 Pro, GPT-5.4, and Kimi K2.6 based agents, 71.4\%--81.8\% of failures fall into Suboptimal datapath design, Suboptimal pipeline design, or Suboptimal data/control-path design.
The correlations support the same view: Timing Closure Rate is negatively correlated with Inefficient STA report-driven optimization (-0.70) and Optimization Flow failures (-0.41), but positively correlated with Suboptimal datapath design (+0.81).

\findingbox{
    \textbf{Findings of RQ2:} The 6 failure categories expose missing capabilities of coding agents in RTL domain knowledge, RTL architecture design, and optimization flow.
    Lower-closure agents mainly fail in report-driven optimization and optimization-flow decisions, while higher-closure agents' remaining failures concentrate in RTL design trade-offs.
}

\section{TicTacSkill: Agent Skill for Timing Closure}
\label{sec:skill}

In this section, we introduce TicTacSkill, a plug-and-play method designed to improve coding agents on ATC tasks.
We first describe how TicTacSkill derives SOPs for timing closure from the failure analysis identified in Section~\ref{sec:analysis}.
We then evaluate this method and show its improvement over the original agent baseline.

\subsection{Skill Design}
\label{sec:skill_design}


Based on the analysis in Section~\ref{sec:analysis}, we propose \textbf{TicTacSkill}, an agent skill that encodes the failure categories identified in Section~\ref{sec:analysis} as Standard Operating Procedures (SOPs) for ATC.
Before solving a task, the coding agent receives this skill that specifies how to inspect timing evidence, localize critical paths, propose RTL rewrites, validate candidates, and decide when to pivot.

\textbf{Critical Path Localization.}
For Inefficient STA report-driven optimization, the skill requires the agent to map slack, startpoint, endpoint, dominant operators, and suspected RTL assignments to the actual critical cone before editing.

\textbf{Datapath Design.}
For Suboptimal datapath design, the skill requires the agent to state which measurable property the candidate changes, such as logic depth, operation width, mux order, fanout, register boundary, or worst-path class.

\textbf{Pipeline Design.}
For Suboptimal pipeline design, the skill requires evidence that a new or moved register actually splits the violating stage rather than only adding latency around the same combinational cone.

\textbf{Control/Decode Placement.}
For Suboptimal data/control-path design, it checks whether flags, comparisons, enables, and state/decode logic are serialized after wide data movement and therefore should be moved earlier or narrowed.

\textbf{Evidence-Driven Search.}
For Unsound exploration, it forces each iteration to compare candidates by correctness, WNS/TNS, violation count, and worst-path evidence, and to abandon repeated edits that do not change the limiter.

\textbf{Pre-PnR/Post-PnR Mismatch.}
For Unsound acceptance, it treats nominal pre-PnR closure as insufficient unless the candidate also has margin or a stated calibration to the final post-PnR objective.

\subsection{Experimental Results}
\label{sec:skill_experiment}

\begin{table}[t]
    \centering
    \caption{TicTacSkill ablation on the three models with skill-enabled runs. Each w/ skill row is paired with the same model and task without the skill.}
    \label{tab:tictacskill}
    \scriptsize
     \renewcommand{\arraystretch}{0.9}
    \begin{tabular}{@{}llrrrr@{}}
        \toprule
        \textbf{Model}            & \textbf{Setting} & \makecell{\textbf{Closure}} & \makecell{\textbf{Avg.}\\\textbf{$\Delta$ WNS\%}} & \makecell{\textbf{Avg.}\\\textbf{$\Delta$ EDDP\%}} & \makecell{\textbf{Avg.}\\\textbf{$\Delta$ ADP\%}} \\
        \midrule
        \multirow{2}{*}{GPT-5.4}  & w/o skill        & 53.3\%                      & 48.7\%                                            & 5.5\%                                              & \textbf{-2.5\%}                                   \\
                                  & w/ skill         & \textbf{56.7\%}             & \textbf{66.4\%}                                   & \textbf{16.3\%}                                    & -3.6\%                                            \\
        \midrule
        \multirow{2}{*}{MiniMax}  & w/o skill        & 23.3\%                      & -1.7\%                                            & \textbf{-0.7\%}                                    & \textbf{-9.6\%}                                   \\
                                  & w/ skill         & \textbf{36.7\%}             & \textbf{17.0\%}                                   & -7.2\%                                             & -10.9\%                                           \\
        \midrule
        \multirow{2}{*}{DeepSeek} & w/o skill        & 33.3\%                      & 44.6\%                                            & 11.6\%                                             & -4.3\%                                            \\
                                  & w/ skill         & \textbf{43.3\%}             & \textbf{62.1\%}                                   & \textbf{14.9\%}                                    & \textbf{-3.0\%}                                   \\
        \bottomrule
    \end{tabular}
\end{table}

We evaluate TicTacSkill by comparing each skill-enabled agent run against the corresponding original run under the same agent and task settings.
Table~\ref{tab:tictacskill} reports the ablation results.
Across all three agents, TicTacSkill improves Timing Closure Rate and Avg. $\Delta$ WNS\%, showing that the failure categories can be converted into useful guidance for timing repair.
The GPT-5.4-based agent improves from 53.3\% to 56.7\% Timing Closure Rate and from 48.7\% to 66.4\% Avg. $\Delta$ WNS\%.
The MiniMax M2.7-based agent shows the largest gain in Timing Closure Rate, increasing from 23.3\% to 36.7\%, while its Avg. $\Delta$ WNS\% moves from -1.7\% to 17.0\%.
The DeepSeek V4 Flash-based agent improves from 33.3\% to 43.3\% Timing Closure Rate and from 44.6\% to 62.1\% Avg. $\Delta$ WNS\%.

\findingbox{
    \textbf{Findings of RQ3:} Targeting the observed failure categories with explicit timing-closure SOPs improves agents' Timing Closure Rate and WNS improvement on the tested agents.
    This shows that the TicTacSkill pattern of turning trajectory-derived failure analysis into a reusable agent skill is effective for improving agents' timing-closure capability.
}

\section{Discussion and Limitations}
\label{sec:discussion}

TicTacBench is an initial benchmark with 30 timing-closure tasks.
This scale is sufficient for exposing clear capability gaps and recurring failure mechanisms, but in future versions we will expand the task set to improve statistical coverage and support finer-grained comparisons across model families.
The current task setting is also focused on single-module combinational designs and single-clock sequential designs.
This scope keeps the benchmark controllable and reproducible, but it does not yet cover the broader range of RTL timing-closure cases found in larger designs.
In future benchmark extensions, we will include more diverse module structures, timing bottlenecks, and design patterns so that ATC evaluation can measure agent capability across a wider task distribution.

\section{Conclusion}
\label{sec:conclusion}

This paper introduces Agent-based Timing Closure as an evaluation target for coding agents that must repair RTL using timing evidence and are judged by post-PnR setup closure.
Across more than 300 runs of coding agents driven by 8 frontier LLMs, the best agent closes only 53.3\% of tasks, showing that current agents still struggle to achieve real timing closure.
TicTacBench can serve as a foundation for moving RTL agents from generating correct modules toward solving implementation-constrained hardware optimization tasks.

\balance
\bibliographystyle{IEEEtran}
\bibliography{main}

@misc{pinckneyRevisitingVerilogEvalYear2025,
  title         = {Revisiting {{VerilogEval}}: {{A Year}} of {{Improvements}} in {{Large-Language Models}} for {{Hardware Code Generation}}},
  shorttitle    = {Revisiting {{VerilogEval}}},
  author        = {Pinckney, Nathaniel and others},
  year          = 2025,
  month         = feb,
  number        = {arXiv:2408.11053},
  eprint        = {2408.11053},
  primaryclass  = {cs},
  publisher     = {arXiv},
  doi           = {10.48550/arXiv.2408.11053},
  urldate       = {2026-04-05},
  archiveprefix = {arXiv}
}

@inproceedings{swebench2024,
  title     = {SWE-bench: Can Language Models Resolve Real-world GitHub Issues?},
  author    = {Jimenez, Carlos E. and others},
  year      = {2024},
  booktitle = {International Conference on Learning Representations},
  url       = {https://openreview.net/forum?id=VTF8yNQM66}
}

@misc{rtlopt2026,
  title   = {A New Benchmark for the Appropriate Evaluation of RTL Code Optimization},
  author  = {Lu, Yao and others},
  year    = {2026},
  journal = {CoRR},
  volume  = {abs/2601.01765},
  doi     = {10.48550/arXiv.2601.01765},
}

@misc{verilogeval2023,
  title   = {VerilogEval: Evaluating Large Language Models for Verilog Code Generation},
  author  = {Liu, Mingjie and others},
  year    = {2023},
  journal = {CoRR},
  volume  = {abs/2309.07544},
  doi     = {10.48550/arXiv.2309.07544},
}

@article{rtlcoder2025,
  title   = {RTLCoder: Fully Open-Source and Efficient LLM-Assisted RTL Code Generation Technique},
  author  = {Liu, Shang and others},
  year    = {2025},
  journal = {IEEE Transactions on Computer-Aided Design of Integrated Circuits and Systems},
  volume  = {44},
  number  = {4},
  pages   = {1448--1461},
  doi     = {10.1109/TCAD.2024.3483089},
}

@misc{rtlrepo2024,
  title   = {RTL-Repo: A Benchmark for Evaluating LLMs on Large-Scale RTL Design Projects},
  author  = {Allam, Ahmed and others},
  year    = {2024},
  journal = {CoRR},
  volume  = {abs/2405.17378},
  doi     = {10.48550/arXiv.2405.17378},
  url     = {https://ieeexplore.ieee.org/document/11310982}
}

@inproceedings{rtlrewriter2024,
  title     = {RTLRewriter: Methodologies for Large Models aided RTL Code Optimization},
  author    = {Yao, Xufeng and others},
  year      = {2024},
  booktitle = {IEEE/ACM International Conference on Computer-Aided Design},
  pages     = {98:1--98:7},
  doi       = {10.1145/3676536.3676775},
}

@inproceedings{openllmrtl2025,
  title     = {OpenLLM-RTL: Open Dataset and Benchmark for LLM-Aided Design RTL Generation},
  author    = {Liu, Shang and others},
  year      = {2025},
  booktitle = {Proceedings of the 43rd IEEE/ACM International Conference on Computer-Aided Design},
  pages     = {60:1--60:9},
  doi       = {10.1145/3676536.3697118},
}

@misc{cvdp2025,
  title   = {Comprehensive Verilog Design Problems: A Next-Generation Benchmark Dataset for Evaluating Large Language Models and Agents on RTL Design and Verification},
  author  = {Pinckney, Nathaniel and others},
  year    = {2025},
  journal = {CoRR},
  volume  = {abs/2506.14074},
  doi     = {10.48550/arXiv.2506.14074},
}

@misc{symrtlo2025,
  title   = {SymRTLO: Enhancing RTL Code Optimization with LLMs and Neuron-Inspired Symbolic Reasoning},
  author  = {Wang, Yiting and others},
  year    = {2025},
  journal = {CoRR},
  volume  = {abs/2504.10369},
  doi     = {10.48550/arXiv.2504.10369},
}

@misc{vitad2025,
  title   = {ViTAD: Timing Violation-Aware Debugging of RTL Code using Large Language Models},
  author  = {Lv, Wenhao and others},
  year    = {2025},
  journal = {CoRR},
  volume  = {abs/2508.13257},
  doi     = {10.48550/arXiv.2508.13257},
}

@misc{gpt54_2026,
  title  = {Introducing GPT-5.4},
  author = {{OpenAI}},
  year   = {2026},
  url    = {https://openai.com/index/introducing-gpt-5-4/},
  note   = {Published: 2026-03-05, Accessed: 2026-03-30}
}

@misc{claude_sonnet46_2026,
  title  = {Introducing Claude Sonnet 4.6},
  author = {{Anthropic}},
  year   = {2026},
  url    = {https://www.anthropic.com/news/claude-sonnet-4-6},
  note   = {Published: 2026-02-17, Accessed: 2026-03-30}
}

@misc{kimi_k26_2026,
  title         = {Kimi K2: Open Agentic Intelligence},
  author        = {{Moonshot AI}},
  year          = {2025},
  eprint        = {2507.20534},
  archiveprefix = {arXiv},
  primaryclass  = {cs.CL}
}

@techreport{deepseek_v4_2026,
  title       = {DeepSeek-V4: Towards Highly Efficient Million-Token Context Intelligence},
  author      = {{DeepSeek-AI}},
  year        = {2026},
  institution = {DeepSeek-AI},
  note        = {Technical report}
}

@misc{minimax_m27_2026,
  title         = {The MiniMax-M2 Series: Mini Activations Unleashing Max Real-World Intelligence},
  author        = {{MiniMax}},
  year          = {2026},
  eprint        = {2605.26494},
  archiveprefix = {arXiv},
  primaryclass  = {cs.CL}
}

@misc{glm51_fp8_2026,
  title         = {GLM-5: from Vibe Coding to Agentic Engineering},
  author        = {{GLM-5-Team} and others},
  year          = {2026},
  eprint        = {2602.15763},
  archiveprefix = {arXiv},
  primaryclass  = {cs.LG}
}

@misc{qwen35_122b_2026,
  title         = {Qwen3 Technical Report},
  author        = {Yang, An and others},
  year          = {2025},
  eprint        = {2505.09388},
  archiveprefix = {arXiv},
  primaryclass  = {cs.CL}
}

@inproceedings{iraware2024,
  title     = {IR-Aware ECO Timing Optimization Using Reinforcement Learning},
  author    = {Jiang, Wenjing and others},
  year      = {2024},
  booktitle = {Proceedings of the ACM/IEEE Workshop on Machine Learning for CAD},
  pages     = {7},
  doi       = {10.1145/3670474.3685945},
}

@inproceedings{esyn2024,
  title     = {E-Syn: E-Graph Rewriting with Technology-Aware Cost Functions for Logic Synthesis},
  author    = {Chen, Chen and others},
  year      = {2024},
  booktitle = {Proceedings of the 61st ACM/IEEE Design Automation Conference},
  pages     = {124:1--124:6},
  doi       = {10.1145/3649329.3656246},
}

@article{retiming1991,
  title   = {Retiming Synchronous Circuitry},
  author  = {Leiserson, Charles E. and others},
  year    = {1991},
  journal = {Algorithmica},
  volume  = {6},
  number  = {1},
  pages   = {5--35},
  doi     = {10.1007/BF01759032},
}

@article{stan2003poweraware,
  title   = {Power-Aware Computing},
  author  = {Stan, Mircea R. and Skadron, Kevin},
  year    = {2003},
  journal = {Computer},
  volume  = {36},
  number  = {12},
  pages   = {35--38},
  doi     = {10.1109/MC.2003.1250889},
}

@misc{invictus2023,
  title   = {INVICTUS: Optimizing Boolean Logic Circuit Synthesis via Synergistic Learning and Search},
  author  = {Chowdhury, Animesh Basak and others},
  year    = {2023},
  journal = {CoRR},
  volume  = {abs/2305.13164},
  doi     = {10.48550/arXiv.2305.13164},
}

@inbook{kahng2011,
  title     = {Timing Closure},
  author    = {Kahng, Andrew B. and others},
  year      = {2011},
  booktitle = {VLSI Physical Design: From Graph Partitioning to Timing Closure},
  pages     = {219--264},
  doi       = {10.1007/978-90-481-9591-6_8},
  isbn      = {9789048195916},
  publisher = {Springer Netherlands}
}

@article{clocktreeaware2016,
  title   = {Clock-Tree-Aware Incremental Timing-Driven Placement},
  author  = {Livramento, Vinicius S. and others},
  year    = {2016},
  journal = {ACM Transactions on Design Automation of Electronic Systems},
  doi     = {10.1145/2858793},
}

@inproceedings{openlane2020,
  author    = {Shalan, Mohamed and Edwards, Tim},
  booktitle = {2020 IEEE/ACM International Conference On Computer Aided Design (ICCAD)},
  title     = {Building OpenLANE: A 130nm OpenROAD-based Tapeout-Proven Flow: Invited Paper},
  year      = {2020},
  pages     = {1--6},
  doi       = {10.1145/3400302.3415735},
}

@inproceedings{rtlbench2025,
  author    = {Zhigang Fang and
               Renzhi Chen and
               Yang Guo and
               Huadong Dai and
               Lei Wang},
  title     = {RTLBench: {A} Multi-Dimensional Benchmark Suite for Evaluating LLM-Generated
               {RTL} Code},
  booktitle = {43rd {IEEE} International Conference on Computer Design, {ICCD} 2025,
               Richardson, TX, USA, November 10-12, 2025},
  pages     = {566--573},
  publisher = {{IEEE}},
  year      = {2025},
  doi       = {10.1109/ICCD65941.2025.00087},
  bibsource = {dblp computer science bibliography, https://dblp.org}
}



\end{document}